\documentclass[letterpaper, 10 pt, conference]{ieeeconf}  

\usepackage{graphicx}
\usepackage{float} 
\usepackage{amsmath}
\usepackage{amssymb}
\usepackage{balance}
\usepackage{subcaption}
\usepackage{hyperref}
\IEEEoverridecommandlockouts                              

\title{\LARGE \bf
Revisiting Space Mission Planning: A Reinforcement Learning-Guided Approach for Multi-Debris Rendezvous
}

\author{Agni Bandyopadhyay$^{1}$ and Günther Waxenegger-Wilfing$^{2}$
\thanks{$^{1}$Agni Bandyopadhyay is pursuing his Doctoral degree with the Faculty of Mathematics and Computer Science,
        Julius-Maximilians-Universität Würzburg, Sanderring 2, 97070 Würzburg, Germany
        {\tt\small agni.bandyopadhyay@uni-wuerzburg.de}}%
\thanks{$^{2}$Günther Waxenegger-Wilfing is a Professor with the Faculty of Mathematics and Computer Science,
        Julius-Maximilians-Universität Würzburg, Sanderring 2, 97070 Würzburg, Germany
        {\tt\small guenther.waxenegger@uni-wuerzburg.de}}%
}

\begin{document}

\maketitle

\begin{abstract}
This research introduces a novel application of a masked Proximal Policy Optimization (PPO) algorithm from the field of deep reinforcement learning (RL), for determining the most efficient sequence of space debris visitation, utilizing the Lambert solver as per Izzo's adaptation for individual rendezvous. The aim is to optimize the sequence in which all the given debris should be visited to get the least total time for rendezvous for the entire mission. A neural network (NN) policy is developed, trained on simulated space missions with varying debris fields. After training, the neural network calculates approximately optimal paths using Izzo's adaptation of Lambert maneuvers. Performance is evaluated against standard heuristics in mission planning. The reinforcement learning approach demonstrates a significant improvement in planning efficiency by optimizing the sequence for debris rendezvous, reducing the total mission time by an average of approximately {10.96\%} and {13.66\%} compared to the Genetic and Greedy algorithms, respectively. The model on average identifies the most time-efficient sequence for debris visitation across various simulated scenarios with the fastest computational speed. This approach signifies a step forward in enhancing mission planning strategies for space debris clearance.

\end{abstract}

\section{INTRODUCTION}
Space debris, commonly referred to as space junk, is any non-functional, artificial material orbiting the Earth. This debris predominantly accumulates in low Earth orbits, but significant quantities are also found near and above geostationary orbits. The European Space Agency's statistical model \cite{master} estimates the presence of approximately 36,500 space debris objects larger than 10 cm, over a million objects ranging from 1 cm to 10 cm, and around 130 million objects measuring 1 mm to 1 cm in size \cite{esa2023}. A notable incident occurred during the STS-7 mission in 1983, when a paint fleck of merely 0.2 mm struck the shuttle's window, creating a 0.4 mm deep pit. This event, though seemingly minor, exceeded the damage threshold for reusing the window's outer pane in future missions and stands as the first recorded instance of Space Shuttle damage caused by orbital debris \cite{nasa-handbook}. The Kessler Syndrome \cite{stenger2002}, \cite{olson1998} highlights the risk of a cascading effect, where increased debris density could lead to further debris generation. This phenomenon poses a significant threat to future space activities in these debris-laden orbits, as emphasized in the recent report by NASA \cite{nasa2023}. The report underscores the urgent need for enhanced debris mitigation efforts, more than ever before in our space exploration history.
However, as with every other space mission, mission planning comprises a crucial part. Thus an optimised mission planning can help with respect to fuel efficiency or in optimising the total time for rendezvousing with all given debris, which is the focus of our paper. 

\begin{figure}[H]
    \includegraphics[width=0.9\linewidth]{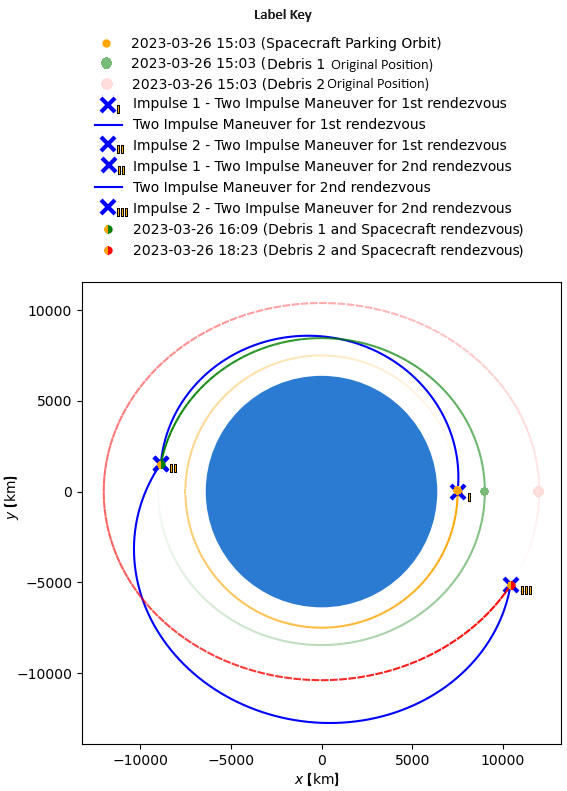}
    \caption{An example problem where two debris rendezvous are conducted. Our spacecraft uses $X_I$ to rendezvous with the first debris. $X_{II}$ represents both the impulses applied on the Spacecraft at the same instant one to rendezvous with Debris 1 and the other one to start the next rendezvous maneuver for Debris 2. $X_{III}$ is the retardation impulse applied to our spacecraft to rendezvous with Debris 2.}
    \label{fig:lamberttwo}
\end{figure}

This research contributes to filling this gap by introducing an improved model for mission planning that optimizes schedules to enhance debris clearance efficiency. Focusing the order in which the debris should be visited so that all of them are rendezvoused in the fastest possible time. By leveraging advanced algorithms in machine learning, specifically tailored to the unique dynamics of space debris, our approach not only predicts time efficient debris rendezvous sequences but does so faster than the traditional heuristics that have been implemented for similar problems. Such advancements represent a significant step forward in debris mitigation technologies.

Our study provides a comprehensive and scalable solution that can be adapted for various types of debris and orbital patterns, offering a robust framework for future space mission planning. This paper will detail the methodologies employed, the results of simulation testing, and the implications for future debris removal strategies.

\section{DEBRIS RENDEZVOUS FRAMEWORK}
\subsection{Active Debris Removal}
Debris removal methods can be broadly classified into two separate groups active and passive removal methods. Active debris removal \cite{adr2013} is the method of removing debris from orbits by first rendezvousing with them and then using active tools like harpoons, robotic arms and others. It is generally employed near the medium Earth orbits where there is no graveyard orbit and the possibility of reentry of the debris into the Earth's atmosphere is low.

\subsection{Travelling Salesman Problem formulation for Active Debris Removal}
Mission planning for active debris removal can be cast as Travelling Salesman Problem \cite{tsp2019}. Assuming a spacecraft in a base orbit, one tries to find a path for rendezvousing with all the given
debris within the least amount of time. An illustration for two debris rendezvous is shown in [Fig.\ref{fig:lamberttwo}]. Different optimization algorithms (\cite{ieeesource}, \cite{greedy2009} and \cite{genetic2019}) have been investigated for solving the resulting TSP variant.

\subsection{Lambert's Problem}
\begin{figure}[H]
    \includegraphics[width=1\linewidth]{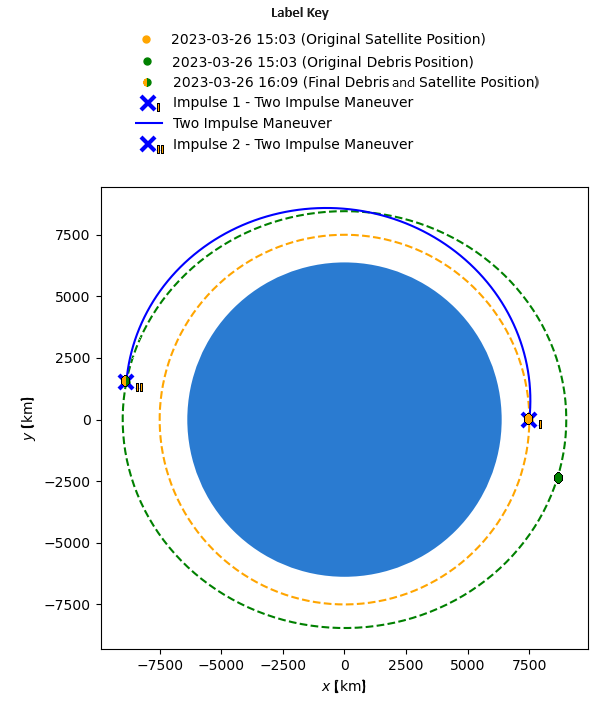}
    \caption{A classical two-impulse debris rendezvous maneuver is demonstrated. It comprises two impulses one for entering the transfer orbit($X_I$) and the other to complete the rendezvous and stay in orbit with the debris($X_{II}$). The point of rendezvous is represented by a multi-coloured orb as the rendezvous means that the spacecraft/satellite occupy the same space in two dimensional representation.}
    \label{fig:lambert}
\end{figure}

For rendezvousing with the debris using the spacecraft, we use the modified Lambert's problem (or Izzo's adaptation of Lambert problem) \cite{izzolambert} to express and then solve for the time of flight equation. This modified algorithm by Izzo for solving the Lambert problem is approximately 1.25 times faster to execute (when multiple revolutions are not considered) than the traditionally used Gooding's algorithm. Here our spacecraft uses only two impulses: once at the start to set the trajectory and finally one at the rendezvous point to stay in the same orbit as the target [Fig.\ref{fig:lambert}]. Expensive maneuvers like inclination change are included in our framework and we assume that we always have enough fuel for the complete rendezvous.

\section{TRADITIONAL HEURISTICS}
\subsection{Greedy method}
A greedy algorithm \cite{greedy2009} functions by choosing the current local optimal solution and hopes to achieve a global optimal solution. It is used because it is one of the fastest methods to get a good solution, which might not be the best solution. With reference to our modified travelling salesman problem \cite{grredytsp}, at every move it chooses the debris which takes the least time to rendezvous and progresses forward.

\subsection{Genetic algorithm}
Genetic algorithms belong to the class of evolutionary heuristics \cite{genetic1989}. For the improvement of a population of solutions evolution-inspired operators are used, comprising three main operations: selection, mutation and crossover \cite{genetic2019}. Genetic algorithms have proved effective with various versions of the travelling salesman problem \cite{gentiong2007comparison}, \cite{genwiak2008intelligent}, \cite{genarora2016better}. For crossover, we used an ordered crossover where holes are generated in a parent chromosome and are filled with attributes from the other parent's chromosomes \cite{genetic1989}. For the mutation we use the shuffle and flip operators \cite{DEAPDocumentation}. The shuffle operator shuffles all attributes of a single chromosome and returns it as a new individual, while the flip operator flips the order of the attributes in the initial chromosome to create a new one.

\section{Reinforcement Learning}

\begin{figure}[H]
    \includegraphics[width=1\linewidth]{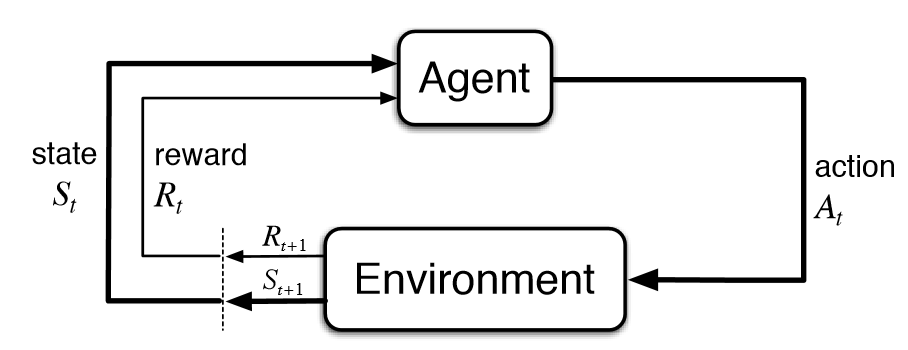}
    \caption{Flowchart of a typical RL algorithm \cite{sharma2020reinforcement}. This demonstrates how an agent in an RL algorithm learns over time how its actions affects the overall environment and learns to adapt over time to maximize the reward.}
    \label{fig:flowchart}
\end{figure}

Reinforcement Learning (RL) \cite{sutton2014reinforcement}, is an area of machine learning where an agent learns to improve its actions by interacting with its environment. The process involves the agent observing the current state of the environment ($S_t$), choosing an action ($A_t$), and receiving feedback in the form of a scalar reward ($R_t$) as represented in Fig.\ref{fig:flowchart} by Sharma \cite{sharma2020reinforcement}, where t is the current time step that is assumed to increase in discrete steps. The objective in RL is to learn a policy for action selection that maximizes cumulative rewards over time, framed within the context of a Markov Decision Process (MDP) \cite{waxeneggerwilfing2021reinforcement}. Deep Reinforcement Learning (DRL) extends RL capabilities using deep neural networks, enabling the agent to handle complex, high-dimensional environments. This approach has found applications in diverse areas (for example: flight control \cite{rlflight} and autonomous spacecraft docking \cite{rldock}), demonstrating its versatility in solving intricate decision-making problems. Thus, the RL framework was chosen due to its generality and ability to construct an amortized optimization solution, in contrast to traditional methods which lack these attributes. An amortized optimization solution refers to the prediction of solutions for optimization problems that share common structures, enhancing the approach's efficacy in addressing the complexity of space debris removal.

\subsection{Proximal Policy Optimization}
Proximal Policy Optimization (PPO) is a model-free, on-policy reinforcement algorithm that aims to optimize policies in a stable and efficient manner. It outperforms other policy gradient methods on various benchmark tests \cite{schulman2017proximal}. Every policy gradient method uses a policy gradient estimator, which is implemented into a stochastic gradient ascent algorithm. PPO is characterized by its clipped surrogate objective function, which helps to stabilize policy updates. The clipped surrogate objective function is given by \cite{schulman2017proximal}:
\begin{align}
\begin{split}
L^{CLIP}(\theta) = \mathbb{E}_t \Biggl[ \min\Biggl(\frac{\pi_{\theta}(a_t|s_t)}{\pi_{\theta_{\text{old}}}(a_t|s_t)} {\mathbb{A}}_t, \\
\quad \text{clip}\Biggl(\frac{\pi_{\theta}(a_t|s_t)}{\pi_{\theta_{\text{old}}}(a_t|s_t)}, 1 - \epsilon, 1 + \epsilon\Biggr)\Biggr) \Biggr]
\end{split}
\end{align}
In this expression (1), $\mathbb{E}_t$ represents the expected value over time steps $t$, $\pi_{\theta}(a_t|s_t)$ is the policy function under parameters $\theta$, giving the probability of taking action $a_t$ given state $s_t$. $\pi_{\theta_{\text{old}}}(a_t|s_t)$ is the policy function under old parameters before the update. ${\mathbb{A}}_t$ denotes the advantage function at time $t$, and $\epsilon$ is a small positive value that defines the clipping range to avoid large policy updates.The 'clip' function in the objective restricts the policy update ratio to between $(1 - \epsilon)$ and $(1 + \epsilon)$, effectively limiting the size of policy updates and promoting gradual learning.

The policy update rule is derived by maximizing the clipped surrogate objective \cite{schulman2017proximal}:

\begin{align}
\begin{split}
\theta_{\text{new}} = \underset{\theta}{\text{argmax}} \Biggl( \mathbb{E}_t \Biggl[ \min\Biggl(\frac{\pi_{\theta}(a_t|s_t)}{\pi_{\theta_{\text{old}}}(a_t|s_t)} {\mathbb{A}}_t, \\
\quad \text{clip}\Biggl(\frac{\pi_{\theta}(a_t|s_t)}{\pi_{\theta_{\text{old}}}(a_t|s_t)}, 1 - \epsilon, 1 + \epsilon\Biggr)\Biggr) \Biggr] \Biggr)
\end{split}
\end{align}
Here, $\theta_{\text{new}}$ represents the updated parameters obtained by maximizing the expectation of the clipped surrogate objective function.

Recent research \cite{ppotsp}, \cite{ppotsp2} has shown that the PPO algorithm works considerably well for solving combinatorial optimization problems. Therefore, we use a modified version of the algorithm for our traveling salesman problem (TSP).

\subsection{Masked Proximal Policy Optimization}
Invalid action masking is a  modification that can be applied to some reinforcement learning algorithms, like to the PPO algorithm \cite{maskppo}. By weeding out invalid actions from the current action space, the algorithm has a much easier task to improve the cumulative reward. Empirical results from previous research \cite{maskppo} have proven that this algorithm works favorably with the scaling of the invalid action space and efficiently trains the algorithm towards meaningful behaviors.

\section{RL ENVIRONMENT AND COMPUTATIONAL SETUP}
When formulating an optimization as an RL problem, the definition of the RL environment including the reward function and the selection of a suitable algorithm are particularly important.

\subsection{Action Space}
The action space encapsulates the range of actions accessible to the agent. In our specific context the action space is discrete, with each action corresponding to the agent's decision regarding the next debris to be targeted.

\subsection{State Space}
The state space must contain all the information that the agent needs to infer the optimal current action. In our case, it is given by an array containing all six Keplerian elements as well as the Cartesian coordinates of the current position for all debris objects, the Cartesian coordinates of the rendezvousing spacecraft and the list of the visited debris objects. In this case, all elements are continuous, except for the list of visited debris objects, which is discrete.
\subsection{Episode Definition and Policy}
An episode in our model is defined as the completion of visits to all debris. The episode length is thus equivalent to the total number of debris. Following Huang et al. (2006) \cite{maskppo}, we incorporate invalid action masking. An invalid action, in our scenario, is defined as any attempt by the interceptor (for example, spacecraft) to revisit a debris site. This mechanism ensures that each piece of debris is visited only once.

\subsection{Reward Function}
The rewards are calculated as follows for all maneuvers except the final rendezvous:

\begin{equation}
{R_t} = -\frac{T_t}{T_{max}}
\end{equation}

where:
\begin{itemize}
\item $R_t$: Reward value.
\item $T_t$: Time to Rendezvous, indicating the actual time taken to complete a specific debris rendezvous.
\item $T_{max}$: Maximum Time for Rendezvous, representing the upper limit for the longest acceptable duration for a rendezvous.
\end{itemize}

In this research, the value of $T_{max}$ is not arbitrary; it is determined based on the maximum expected time for any single rendezvous maneuver within the scope of our simulations. The normalization of the time-to-rendezvous by $T_{max}$ serves a dual purpose: it ensures that the reward remains within a consistent range irrespective of mission duration variations, and it simplifies the comparison of results across different mission scenarios. This formulation ensures that the reward $R_t$ is normalized between -1 and 0, with -1 being the least efficient and 0 the most efficient outcome. 

Our sensitivity analyses have shown that such normalization does not impact the final optimization result, but rather ensures the algorithm's learning process is stable and efficient across diverse operational contexts. For the final rendezvous, we use the same reward calculation but add an additional factor of +1 as further bonus. This approach balances individual maneuver efficiency with the overall mission objective, aligning the algorithm's performance with the goal of planning time-optimal multi-debris rendezvous.

\subsection{Dynamic Decision-Making of the Agent}
Unlike traditional methods where the entire debris visitation sequence is pre-planned, our agent adopts a dynamic decision-making approach \cite{maqbool}. At each step, the agent visits one debris object and then analyzes the current scenario to determine the next target. This method offers significant flexibility for real-time adjustments, such as collision avoidance or removal operations that last longer than expected. Consequently, the agent's ability to adapt its path on-the-fly increases the overall efficiency of the debris clearing operation.

\section{Overview of Simulation Test Case and Algorithm Configuration}
Using a MacBook Pro with 64GB of memory and M1 processor, along with Python 3.10, we employed physics simulation libraries such as poliastro \cite{poliastro} and astropy \cite{astropy2013}, as well as optimization algorithms implemented via DEAP \cite{deap2012} and Stable Baselines3 \cite{stable-baselines3}. The Iridium 33 debris data, sourced from Celestrak \cite{celestrak}, was used for simulations within the period from November 23 to November 26, 2023. We divided the dataset into training (70\%), testing (15\%), and evaluation (15\%) subsets. Each simulation involved selecting ten random debris pieces from the dataset for a given date. Our spacecraft's goal is to rendezvous with all selected debris in the shortest possible time from a given parking orbit, assuming sufficient fuel availability. For detailed computational setup and hyper parameter configurations, refer to Appendix A.

\section{RESULTS AND DISCUSSION}

In this section, we delve into the performance evaluation of our approach, focusing on their efficiency and efficacy in planning and multi-debris rendezvous. Through comparative analysis, we aim to underscore the distinct advantages offered by the RL approach.

\begin{figure}[H]
\includegraphics[width=1\linewidth]{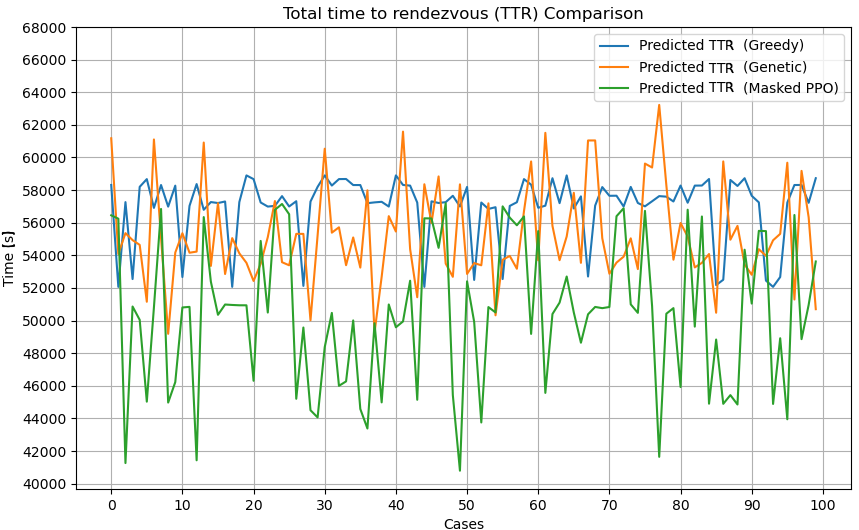}
\caption{Predicted total time to rendezvous (TTR) for all algorithms for all the test cases}
\label{fig:timecomp}
\end{figure}

The performance comparison for the evaluation dataset, as illustrated in Fig. \ref{fig:timecomp}, accentuates the Masked PPO algorithm's capability to outperform traditional algorithms. This efficiency stems from its sophisticated assessment of actions within its operational environment, fostered through extensive training phases. In the evaluation of our algorithm's performance, a significant emphasis was placed on its efficiency and efficacy in planning and orchestrating multi-debris rendezvous missions.\\

\begin{figure}[H]
\centering
\includegraphics[width=\linewidth]{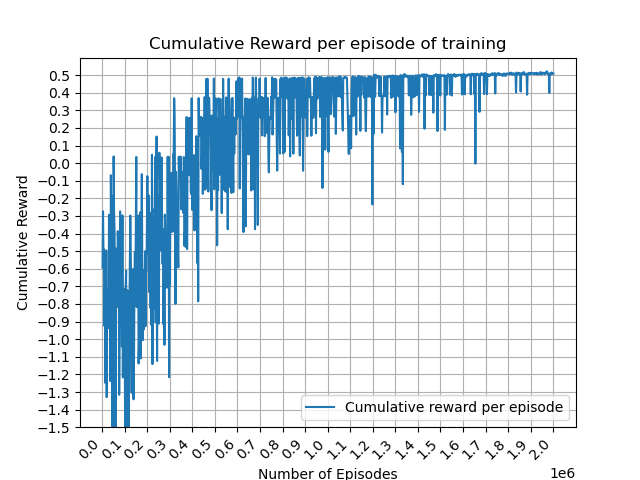}
\caption{Cumulative reward per episode for the Masked PPO algorithm (from Tensorboard log data) is shown here. The goal is to increase this reward over time but also strive towards a deterministic value at the end.}
\label{fig:cumrew}
\end{figure}

Figure \ref{fig:cumrew} charts the learning curve of the Masked PPO algorithm, demonstrating an initial exploratory phase followed by an optimization of the cumulative reward. When an episode is reset, a set of debris objects is selected at random from the training dataset, i.e. the task is randomized. This process significantly improves the agent's understanding of the complex scenario and enables the generalization visible in the evaluation set.

\begin{table}[H]
\centering
\caption{A statistical analysis comparing the predicted total time for rendezvous (TTR) of Genetic, Greedy, and PPO algorithms. The analysis indicates a statistically significant difference in the mean predicted total time for rendezvous (TTR), underscoring the efficiency of the PPO algorithm.}
\label{tab:anova}
\resizebox{\linewidth}{!}{%
\begin{tabular}{|l|c|r|r|r|}
\hline
\textbf{Groups} & \textbf{Count} & \textbf{Sum (in seconds)} & \textbf{Average (in seconds)} & \textbf{Variance (in seconds)} \\ \hline
Genetic TTR & 100 & 5523867.791 & 55238.67791 & 8753908.178 \\ \hline
Greedy TTR & 100 & 5697144.916 & 56971.44916 & 3872476.734 \\ \hline
PPO TTR & 100 & 4918592.247 & 49185.92247 & 56126608.79 \\ \hline
\end{tabular}%
}
\end{table}

A simple statistical analysis was conducted to compare the performance in predicted total time for rendezvous (TTR) between the Genetic, Greedy, and our Masked PPO algorithms, with the results presented in Figure \ref{tab:anova}. The PPO algorithm demonstrated a reduction in average predicted total time for rendezvous (TTR) by approximately {10.96\%} and {13.66\%} compared to the Genetic and Greedy algorithms, respectively. This reflects the potential of our algorithm to optimize space mission planning significantly.

\begin{table}[H]
\centering
\caption{A statistical analysis comparing the execution times of Genetic, Greedy, and PPO algorithms. The analysis indicates a statistically significant difference in the mean execution (or computational) times, underscoring the efficiency of the PPO algorithm.}
\label{tab:anova2}
\resizebox{\linewidth}{!}{%
\begin{tabular}{|l|c|r|r|r|}
\hline
\textbf{Groups} & \textbf{Count} & \textbf{Sum (in seconds)} & \textbf{Average (in seconds)} & \textbf{Variance (in seconds)} \\ \hline
Genetic Execution Time & 100 & 54662.54091 & 546.6254091 & 1117.780564 \\ \hline
Greedy Execution Time & 100 & 32.56012344 & 0.3256012344 & 0.0006626908445 \\ \hline
PPO Execution Time & 100 & 13.4871645 & 0.134871645 & 0.0001851706585 \\ \hline
\end{tabular}%
}
\end{table}

\begin{figure}[H]
    \centering
    \begin{subfigure}{0.45\textwidth} 
        \centering
        \includegraphics[width=\textwidth]{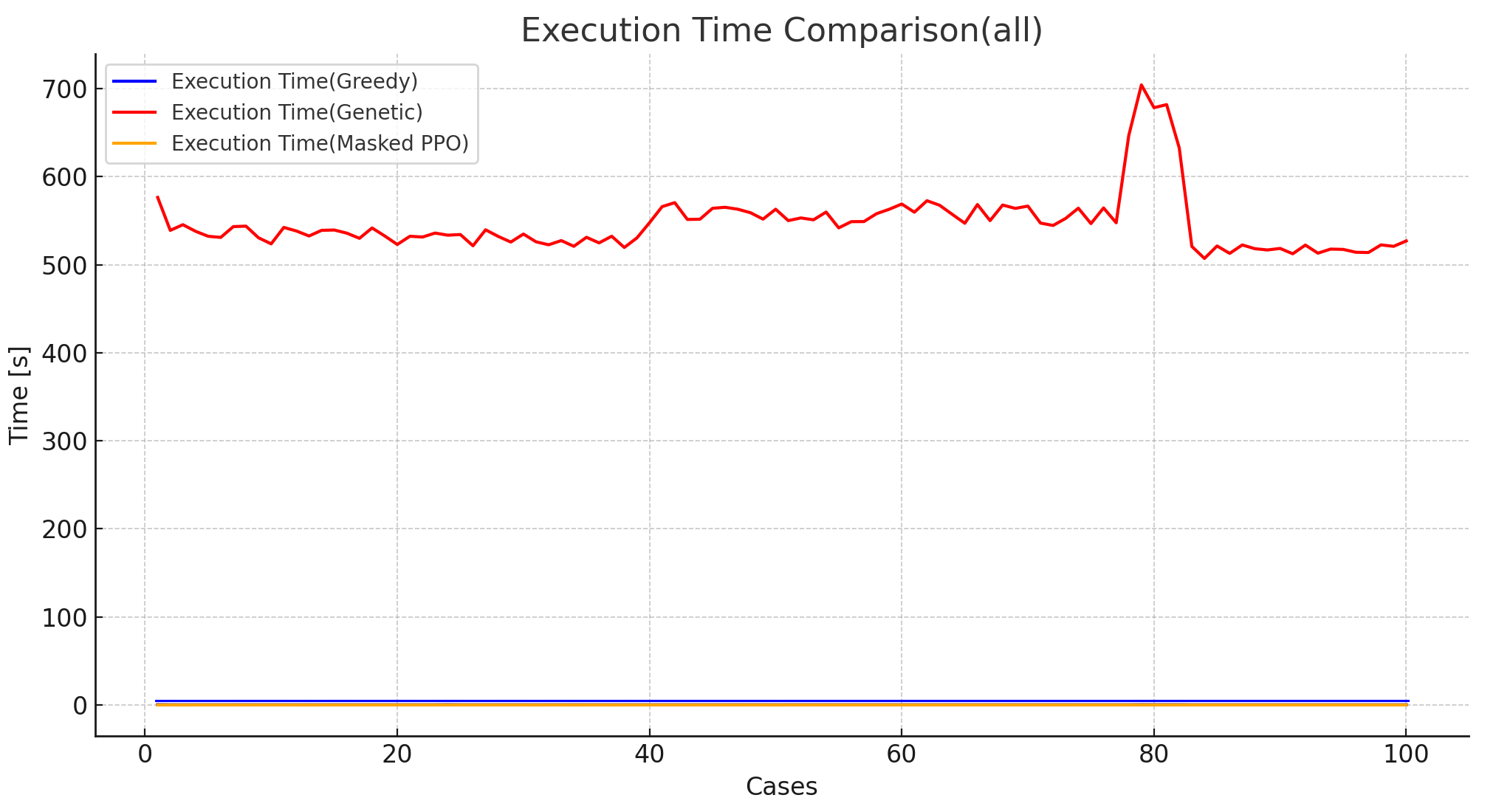} 
        \caption{Comparison of Greedy, Genetic and Masked PPO model's execution time or the time taken to predict the complete order in which the debris should be visited. The execution time series of the Greedy algorithm cannot be seen as it is obscured by the masked PPO model's data. This figure is just to represent that the execution times of the Genetic algorithm is much higher in comparison with the other two.}
        \label{subfig:greedy_genetic_ppo}
    \end{subfigure}
    \hfill 
    \begin{subfigure}{0.45\textwidth} 
        \centering
        \includegraphics[width=\textwidth]{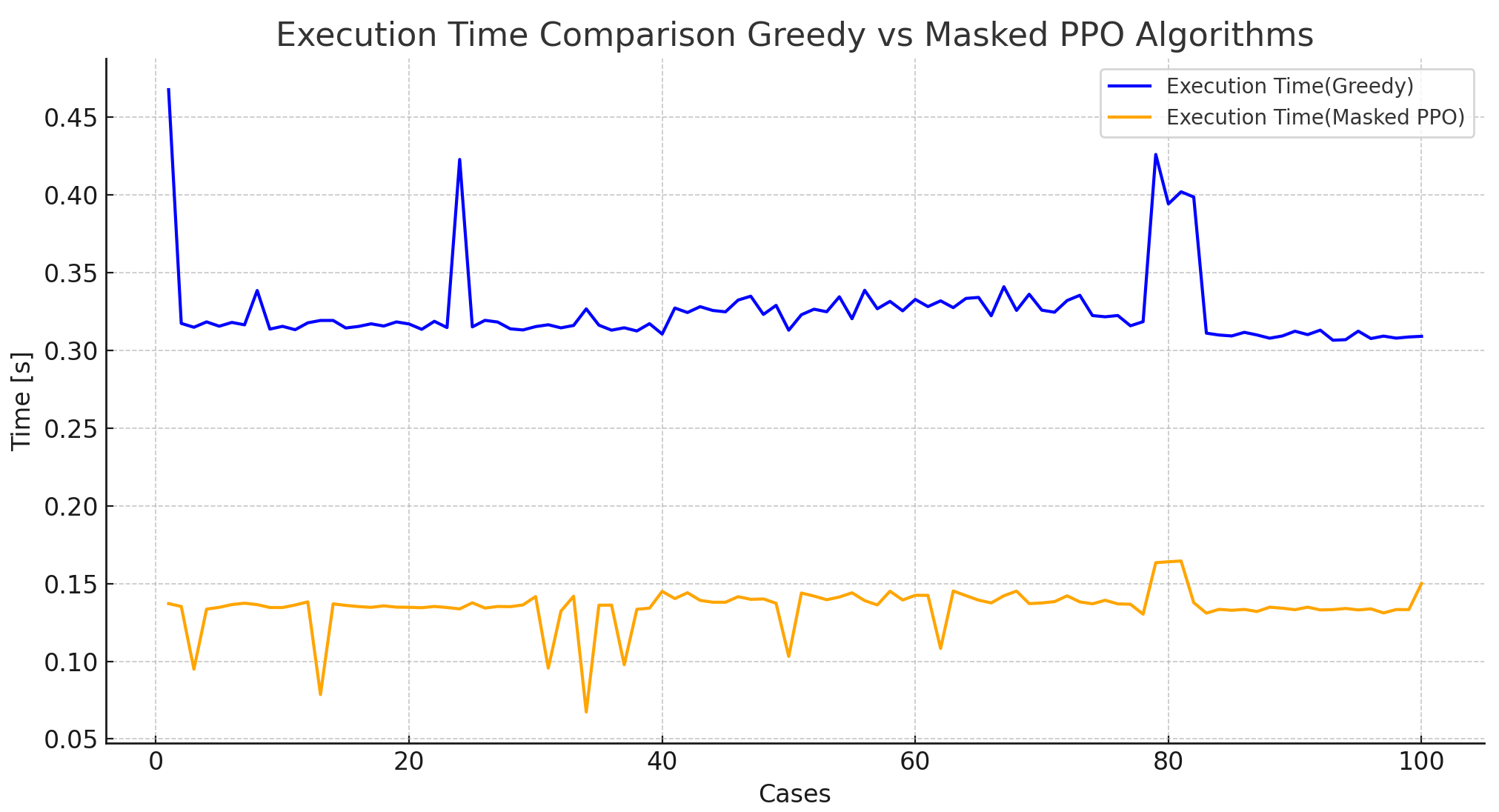} 
        \caption{Comparison of Greedy and Masked PPO model's execution time or the time taken to predict the complete order in which the debris should be visited. This figure is a zoomed in version of the figure above to demonstrate the Greedy and PPO model's execution time as they are of comparable range.}
        \label{subfig:greedy_ppo}
         \end{subfigure}
    \caption{Execution time comparisons highlighting the efficiency of the Masked PPO model against the Greedy and Genetic algorithms.}
    \label{fig:execution_time_comparisons}
\end{figure}

The execution time plots \ref{subfig:greedy_ppo} and \ref{subfig:greedy_genetic_ppo} further elucidates the advantages of employing the Masked PPO algorithm over its counterparts. The execution time analysis depicted in \ref{tab:anova2} illustrate that the Masked PPO algorithm's average execution time is consistently lower than that of the greedy and genetic algorithm. The differences between the algorithms execution time are likely to be exacerbated when dealing with a larger number of debris objects. Consequently, the Reinforcement Learning solutions are better, yielding not only a shorter total rendezvous time but also a notably quicker computational speed (post an extensive training phase) relative to alternative approaches.

\section{SUMMARY AND OUTLOOK}

Reinforcement Learning (RL) is advancing the frontier of combinatorial optimization, a development that our study reinforces. Beyond addressing intricate problems, RL generates on average optimal solutions, which surpass state of the art heuristics and offers efficiency in execution after the extensive training phase. The sequential approach of generating the solution constitutes an essential feature for the integration of complex maneuvers like collision avoidance as well as for refuelling scenarios.

However, RL is not without its challenges. The extensive training phase, necessary for simulating diverse scenarios, demands considerable time. Moreover, the need for expansive datasets to refine model robustness and applicability persists. Despite these challenges, the benefits of implementing RL in space mission planning seem obvious.

As computational capabilities evolve and datasets grow, the effectiveness and applicability of RL algorithms are expected to expand, paving the way for more autonomous spacecraft operations. We hope that our work represents a first step in this exciting direction.




\section*{APPENDIX}
\balance
\section*{Appendix A: Detailed Maskable PPO Training Hyperparameters}

The configuration of our Maskable Proximal Policy Optimization (PPO) algorithm, implemented using Stable Baselines3 \cite{stable-baselines3}, is tailored for the space debris targeting task. The key hyper parameters are as follows:

\begin{itemize}
    \item \textbf{Learning Rate:} \(3 \times 10^{-4}\), controlling the update rate of the agent's policy.
    \item \textbf{Number of Steps:} \(2048\), defining the number of steps collected before updating the model.
    \item \textbf{Batch Size:} \(64\), the size of the batch used in the optimization process.
    \item \textbf{Number of Episodes:} \(200,000\),  where 1 episode consists of 10 time steps.
    \item \textbf{Discount Factor (\(\gamma\)):} \(0.99\), used in calculating the discounted future rewards.
    \item \textbf{GAE Lambda (\(\lambda\)):} \(0.95\), for the Generalized Advantage Estimator (GAE).
    \item \textbf{Clip Range:} \(0.2\), for the PPO clipping in the policy objective function.
    \item \textbf{Value Function Coefficient (vf\_coef):} \(0.5\), the scaling factor for the value function loss in the total loss calculation.
    \item \textbf{Maximum Gradient Norm (max\_grad\_norm):} \(0.5\), used for gradient clipping.
    \item \textbf{Entropy Coefficient (ent\_coef):} \(0.0\), which adds an entropy bonus to the reward to ensure sufficient exploration.
    \item \textbf{Verbose Level:} Set to \(0\) for minimal output during training.
    \item \textbf{Device:} Set to 'auto', allowing the system to choose the appropriate computation device (CPU or GPU).
\end{itemize}

\section*{Appendix B:Terminology and Definitions}
In this paper, we have adopted specific terms that are pivotal to the understanding of the mission design and planning. Here, we clarify these terms to ensure clarity and avoid any potential ambiguity:

\begin{itemize}
  \item \textbf{Parking Orbit/Base Orbit:} The term 'parking orbit' is used interchangeably with 'base orbit' to refer to the initial orbit where the spacecraft begins its debris removal operations.
  \item \textbf{Execution/Computational time:} This term refers to the total duration required for the algorithm or computational process to complete a task. In the context of space mission simulations, it encompasses the period from the initiation of the debris visitation sequence calculation to the output of the complete path, including all computational steps and processes involved. It is completed when all debris are visited. 
\end{itemize}

\section*{Appendix C: Acronyms and Figures}
\section*{List of Acronyms}
\label{appendix:acronyms}

\begin{description}
    \item[AI] Artificial Intelligence 
    \item[PPO] Proximal Policy Optimization
    \item[STS] Space Transportation System
    \item[TSP] Travelling Salesman Problem 
    \item[RL] Reinforcement Learning
    \item[DEAP] Distributed Evolutionary Algorithms in Python
    \item[ADR] Active Debris Removal
    \item[ESA] European Space Agency
    \item[NASA] National Aeronautics and Space Administration
\end{description}

\listoffigures

\bibliographystyle{ieeetr}
\bibliography{references}
\balance

\end{document}